\documentclass{article}
\usepackage{ijcai18}

\usepackage{times}
\usepackage{soul}
\usepackage{url}
\usepackage[hidelinks]{hyperref}
\usepackage[utf8]{inputenc}
\usepackage[small]{caption}

\usepackage{algorithm}
\usepackage{algorithmic}
\usepackage{microtype}
\usepackage{graphicx}
\usepackage{subfigure}
\usepackage{booktabs} 
\usepackage{hyperref}

\title{Generating Natural Language Explanations for Visual Question Answering \\Using Scene Graphs and Visual Attention}

\author{
Shalini Ghosh\thanks{Work done while the first author was at SRI International.}, \qquad
Giedrius Burachas$^{\dagger}$,\qquad
Arijit Ray$^{\dagger}$,\qquad
Avi Ziskind$^{\dagger}$ \\
$^*$\texttt{firstname.lastname@samsung.com}, \qquad
$^{\dagger}$\texttt{firstname.lastname@sri.com} \\
$^*$Samsung Research, $^{\dagger}$SRI International 
}

\begin{document}

\maketitle

\begin{abstract}
In this paper, we present a novel approach for the task of eXplainable Question Answering (XQA), i.e., generating natural language (NL) explanations for the Visual Question Answering (VQA) problem.  We generate NL explanations comprising of the evidence to support the answer to a question asked to an image using two sources of information: (a) annotations of entities in an image (e.g., object labels, region descriptions, relation phrases) generated from the scene graph of the image, and (b) the attention map generated by a VQA model when answering the question. We show how combining the visual attention map with the NL representation of relevant scene graph entities, carefully selected using a language model, can give reasonable textual explanations without the need of any additional collected data (explanation captions, etc). We run our algorithms on the Visual Genome (VG) dataset and conduct internal user-studies to demonstrate the efficacy of our approach over a strong baseline. We have also released a live web demo showcasing our VQA and textual explanation generation using scene graphs and visual attention.\footnote{https://xai.nautilus.optiputer.net/}
\end{abstract}

\section{Introduction}
\label{sec:intro}

Visual Question Answering (VQA) ~\cite{VQA}, the  task of answering natural language questions on images, has garnered a lot of interest as an AI-complete task. While impressive strides have been made on this task using deep networks \cite{kazemi2017show,DBLP:journals/corr/abs-1708-02711}, they are notorious for being opaque/black-boxed to a non-expert user, thus making it hard to understand when/why it predicts an incorrect answer. There have been attempts to make VQA systems more human-like \cite{ray2016question} and on how they can hold conversations if one desires further questioning \cite{ray2017art} \cite{DBLP:journals/corr/DasKGSYMPB16}. However, VQA/conversational agents still cannot explain in natural language why they made a certain decision. This raises issues of trust and reliability since the user cannot judge when to trust the predictions of the model or not. In this paper, we focus on a natural language solution to the eXplainable Question Answering (XQA) task, the task of explaining the answer that was provided  --- specifically, our goal is to automatically generate a natural language sentence that provides evidence to support the answer predicted by a VQA model. Ideally, such an explanation will help people judge and/or trust the answer provided better.  

\begin{figure}[hbtp]
\begin{center}
\centerline{\includegraphics[width=0.9\columnwidth]{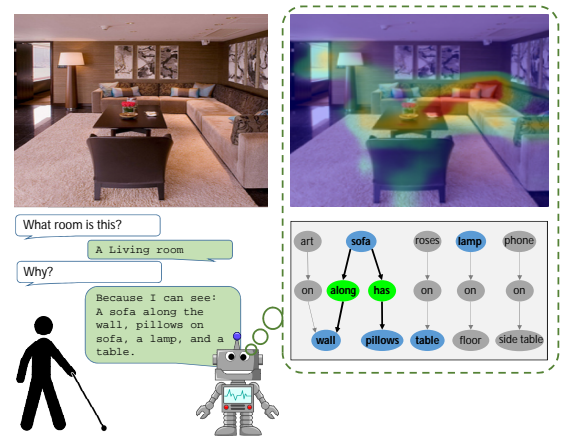}}
\caption{Example Natural Language explanation for an answer to a visually-grounded question. Our approach uses the attention map generated by a visual question-answering model (top right) to identify the relevant components of a scene graph (bottom right) that can be used to provide a justification for the answer given by the model.}
\label{fig:scene_graph}
\end{center}
\end{figure}

Figure \ref{fig:scene_graph} illustrates a summary of our objective. Asked a question, ``What room is this?'', to which the answer is ``living room'', the visually impaired user requests for an explanation in order to ensure the answer is correct. Our Explainable VQA Agent uses the attention heatmap (regions of interest as suggested by the model while answering the question) to pick out relevant information from an annotated image scene graph to generate a natural language explanation phrase --- ``Because I can see: sofa along the wall, pillows on sofa, a lamp, and a table.'' 
Our proposed algorithm uses the visual attention map as a guide to identify the relevant entities from the scene graph (where entities could be objects, relations or descriptions), and then uses NLP techniques using language models to compose a NL representation of those entities to generate a NL explanation for the VQA answer. 
By conducting a small-scale user study, we show evidence that such an approach can generate reasonable textual explanations for answers to questions on images in the Visual Genome Dataset.   




Section~\ref{sec:back} covers the relevant background and some related work. Section~\ref{sec:algo} outlines the core algorithms, while Section~\ref{sec:case} gives some example explanations generated by our algorithm. We performed some initial experiments using the Visual Genome (VG) dataset to demonstrate the effectiveness of our method --- an analysis of those results are described in Section~\ref{sec:exp}. Section~\ref{sec:rel} discusses related approaches, and finally Section~\ref{sec:concl} concludes the paper and gives an overview of possible future work in this area.

\begin{figure}[hbtp]
\begin{center}
\centerline{\includegraphics[width=0.4\columnwidth]{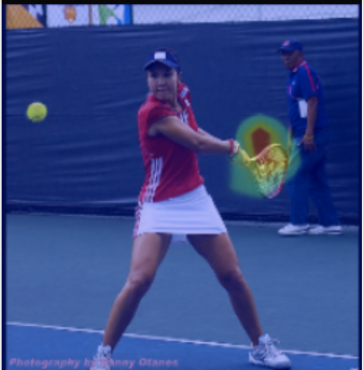}}
\caption{Salient parts of an Visual Genome image highlighted by the attention layers of our VQA Model, corresponding to the question/answer pair ``What is this game? Tennis".}
\label{fig:tennis-b}
\end{center}
\end{figure}
\vspace{-5mm}
\section{Background}
\label{sec:back}

In this section, we briefly describe some of the key models and concepts used in this paper.




\subsection{Visual Question Answering and Attention Layers}

\begin{figure}[hbtp]
\begin{center}
\centerline{\includegraphics[width=0.9\columnwidth]{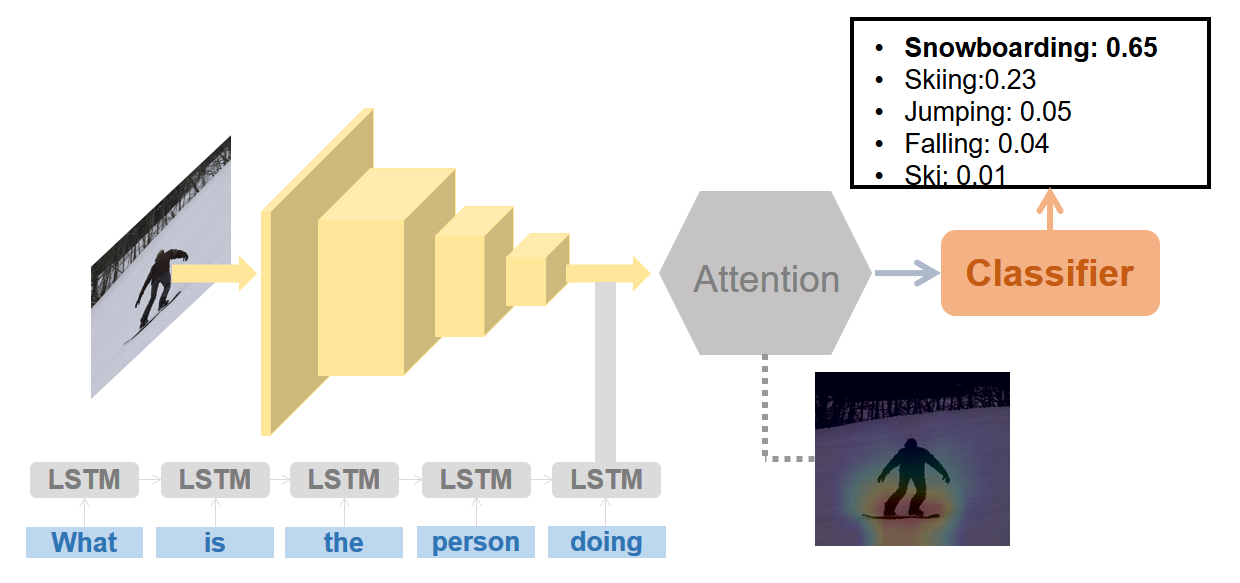}}
\caption{We use a simplistic VQA architecture which is very similar to the SA3 model by Google. The model takes as input a 224 x 224 image and a natural language question (words encoded as 300 dimensional GloVe vectors) and outputs a sigmoidal probability distribution over 3000 answer choices. The attention layer weighs a 7 x 7 x 2048 convolutional map, which is weighted averaged to get a 7 x 7 heatmap. The 7 x 7 heatmap is resized to 224 x 224 to get the attention heatmap for the image.}
\label{fig:VQA}
\end{center}
\end{figure}

Visual Question Answering is the task of answering natural language questions about an image. This requires simultaneous understanding of textual and visual semantics --- deep neural networks have made impressive strides at this task. We use the VQA architecture outlined in Figure~\ref{fig:VQA}. Our model takes as input a 224 x 224 RGB image and a question of at most 15 words. The image is encoded using a ResNet152~\cite{DBLP:journals/corr/HeZRS15} to get a 7x7x2048 image feature representation. The question is encoded using an LSTM which takes in the GloVe~\cite{pennington2014glove} word embeddings of the words, one word at a time. The final LSTM state is used to represent the question features. The attention layer takes in the question and image feature representations and outputs a set of weights to attend on the image features. The weighted image features, concatenated with the question representation, is used to predict the final answer from a set of 3000 answer choices.  

\subsection{Scene Graph}

The scene graph for an image is the graphical representation of its contents, where the nodes are the depicted objects and the edges are the relationships between them (e.g. as shown in Figure~\ref{fig:scene_graph}). Some scene graphs also contain region descriptions (more detailed annotations of an object, or a description of a region containing multiple interacting objects). For example, the Visual Genome data has region descriptions in addition to labeled objects/relations, a sample of which is shown in Figure~\ref{fig:tennis2}. However, most methods for generating scene graphs~\cite{lu:16} generate graphs with only objects and relations, but without region descriptions. Depending on the type of scene graph available, we designed two variants of the explanation candidate generation models: (1) one that generates NL explanations based on region descriptions, and (2) one that generates NL explanations based on objects and relations.


\subsection{Web Language Model}

Given a sequence of words, language models estimate the probability of observing another sequence of words following it. There are various language models available, typically trained on large scale corpora (e.g., Web news data, Wikipedia), which give robust estimates of conditional probabilities of observing one text segment in the context of another text segment~\cite{jozefowicz:16}. We use the Web language model service AzureLM\footnote{https://azure.microsoft.com/en-us/services/cognitive-services/web-language-model/} to get robust estimates of conditional probabilities of text segments.

\section{Algorithm}
\label{sec:algo}

\begin{figure}[hbtp]
\begin{center}
\centerline{\includegraphics[width=\columnwidth]{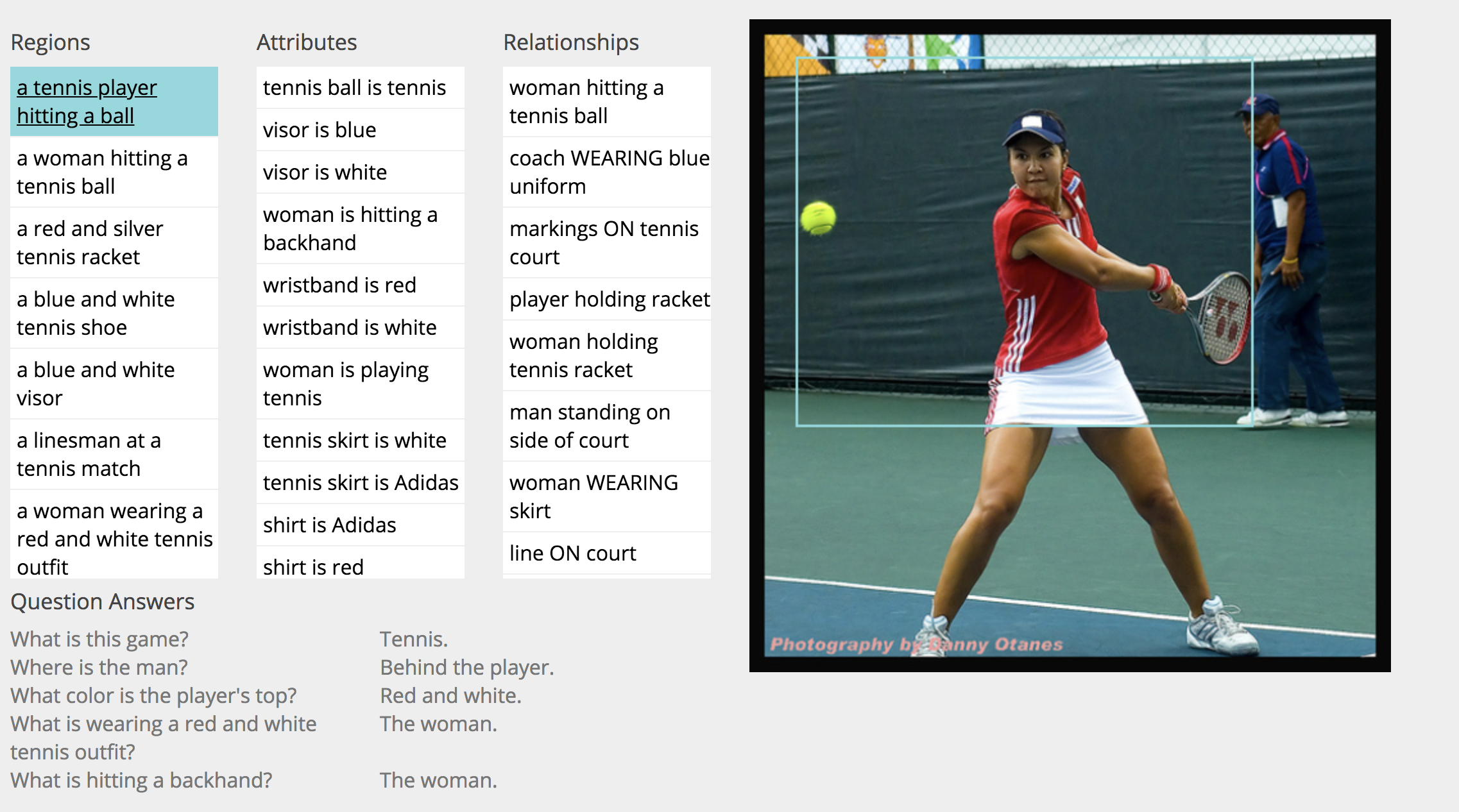}}
\caption{Scene graph of an image from Visual Genome data, showing object attributes, relation phrases and descriptions of regions.}
\label{fig:tennis2}
\end{center}
\end{figure}

One of the key insights in our approach is the observation that the scene graph of an image can be very useful in generating explanations, something that has not been explored in previous work on generation visual explanations. We use the scene graph to retrieve the set of relevant entities for an image, where an entity refers to either an object, relation or region description, and then consider the natural language phrases corresponding to those entities. Note that each entity in the scene graph has an associated bounding box. For example, for the tennis image in Figure~\ref{fig:tennis2}, the bounding box shown in the figure is associated with the region description ``a tennis player hitting a ball". 

For a given image and a question/answer pair, we use the heatmap generated by the visual attention layer to identify the parts of the image that are relevant and retrieve the bounding boxes with a high degree of overlap with these regions. For example, for the attention map shown in Figure~\ref{fig:tennis-b}, corresponding to the question/answer pair ``What is this game? Tennis", the region of the image containing the tennis racket is highlighted, thus identifying this object within the scene graph as relevant for the explanation.  We rank the most important entities from the scene graph using a score function, which includes, among other things: (1) The degree of overlap of the bounding box with the active region of the attention map, and
(2) An estimate of the relevance of the NL representation of the entity w.r.t. the question/answer. 

Our composite score helps us identify entities with NL representations that are deemed relevant to the explanation of the question/answer pair, both from the point of view of the visual attention model as well as the language model. E.g., for Figure~\ref{fig:tennis2}, a region description with high relevance score (from both the visual model and language model) is ``a tennis player hitting a ball" --- we use this high-scoring region description to generate the final explanation ``The picture shows: a tennis player hitting a ball".

\begin{figure}[hbtp]
\begin{center}
\centerline{\includegraphics[width=\columnwidth]{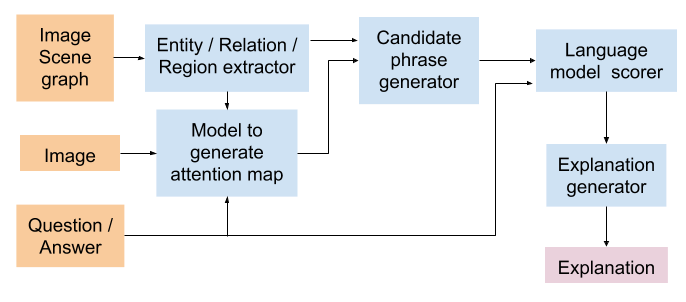}}
\caption{Proposed Explanation Generation Pipeline.}
\label{fig:scoring_pipeline}
\end{center}
\end{figure}

Figure~\ref{fig:scoring_pipeline} shows the overall flow of explanation generation used in our approach, where we generate explanations based on inferences by a language model and the visual attention layer. When we have region descriptions corresponding to an image, along with the visual attention model, we use the approach in Algorithm~\ref{algo1} to generate the explanations. The score function used to rank regions is:

\begin{eqnarray}
\label{eq:score}
score(D, QA) &=& attentionScore(R(D)|QA) \nonumber \\
&\times&\ lmScore(D|QA) \nonumber \\
&\times&\ sqrt(len(D)) \nonumber \\
&\times&\ 1/log(area(R(D)))
\end{eqnarray}

\noindent where D corresponds to a region description, $R(D)$ is the bounding box of that region, $attention(R(D)|QA)$ is the total attention score from the visual attention layer for region R(D) for question/answer Q/A, $lmScore(D|QA)$ is the language model score of text of D in the context of the text of QA, $len(D)$ is the length of description D and $area(R(D))$ is the area of R(D). Given two regions for related areas, this score function selects the region with tighter bounding box and  longer (i.e., richer) description. The slot filler explanation generation in this case simply adds a prefix ``The picture shows:'' to the generated explanation --- in the future, we plan to explore other more complex slot-filter templates (if necessary).

\begin{algorithm}
\caption{Generate XQA explanations using regions}
\begin{algorithmic}
\REQUIRE Image I, question Q, answer A, regions RS in I with descriptions DS, visual (attention) and language (LM) scoring models.
\ENSURE Explanations E for Q/A, sorted by relevance.
\FORALL{Description D in DS, for region R in RS}
\STATE Compute relevance of D to Q/A using score(D, QA) defined in Equation~\ref{eq:score}, using attention and LM models.
\ENDFOR
\STATE D' = Descriptions D sorted by decreasing score(D, QA)
\STATE {\bf return} E = Slot-filler explanations generated from D'
\end{algorithmic}
\label{algo1}
\end{algorithm}

When the scene graph of an image just has objects and relations (and no region descriptions, like in the Visual Genome data), then we use Algorithm~\ref{algo2} to generate the explanations. In this algorithm, we find out the relevant objects/relations and then use a graph traversal algorithm to find a set of connected relations that form a connected component in the graph --- we use the connected component to generate a ``descriptive" explanation (outlined in Algorithm~\ref{algo3}). Thus, when region descriptions are not available, we can instead use these descriptive explanations for the connected component of relations relevant to the question/answer.

\begin{algorithm}[ht]
\caption{Generate XQA explanations using objects/relations}
\begin{algorithmic}
\REQUIRE Image I, question Q, answer A, objects or relations OS in the scene graph, visual (attention) and language (LM) models.
\ENSURE Explanations E for Q/A, sorted by relevance.
\FORALL{Object O in OS}
\STATE Compute relevance of O to Q/A using score(O, QA) defined in Equation~\ref{eq:score}, using attention and LM models.
\ENDFOR
\STATE O' = Objects O sorted by decreasing score(O, QA)
\STATE {\bf return} E = Descriptive explanations generated from O' using Algorithm~\ref{algo3}.
\end{algorithmic}
\label{algo2}
\end{algorithm}

\begin{algorithm}[ht]
\caption{DFSSortedWithEmit(N): Generate descriptive explanations using objects/relations for graph rooted at N}
\begin{algorithmic}
\REQUIRE Graph G = (OS, RS), OS is set of objects (nodes), RS is set of relations (edges), maximum number of objects used in explanation kNumTermsInExplanation, language (LM) model.
\ENSURE Explanations E for Q/A.
\STATE LR = list of relations RS in decreasing order of LM model score
\STATE LO = list of objects OS in decreasing order of LM model score
\STATE Explanation list EL = []
\FORALL{Relation R(O, O’) in LR}
    \IF{O is in LO and O is not marked}
    \STATE Phrase P = DFSSortedWithEmit(O)
    \STATE Add phrase P to EL
    \ENDIF
\ENDFOR
\IF{size(EL) $<$ kNumTermsInExplanation}
    \FORALL{For unmarked object O in LO:}
    \STATE Phrase P’ = DFSSortedWithEmit(O)
    \STATE Add phrase P’ to EL until size(EL) = kNumTermsInExplanation
    \ENDFOR
\ENDIF
\STATE Create explanations E with phrases from list EL using slot-filling
\STATE {\bf return} E
\end{algorithmic}
\label{algo3}
\end{algorithm}

\section{Example Explanations}
\label{sec:case}

In this section, we illustrate our algorithm using the Visual Genome (VG) data, providing a few examples in which the multi-modal explanation generation algorithm performs well. For the tennis image and associated Q/A, Algorithm~\ref{algo1}, the region-based algorithm gave the following results using the multi-modal explanation generation approach:\\

\noindent {\bf Q/A}: What is this game? Tennis. {\bf Explanation}:\\
\noindent 1. The picture shows: a tennis court\\
\noindent 2. The picture shows: a tennis player hitting a ball\\
\noindent 3. The picture shows: a woman hitting a tennis ball\\
\noindent 4. The picture shows: a red and silver tennis racket\\
\noindent 5. The picture shows: a blue and white tennis shoe\\

\begin{figure}[hbtp]
\begin{center}
\centerline{\includegraphics[width=0.75\columnwidth]{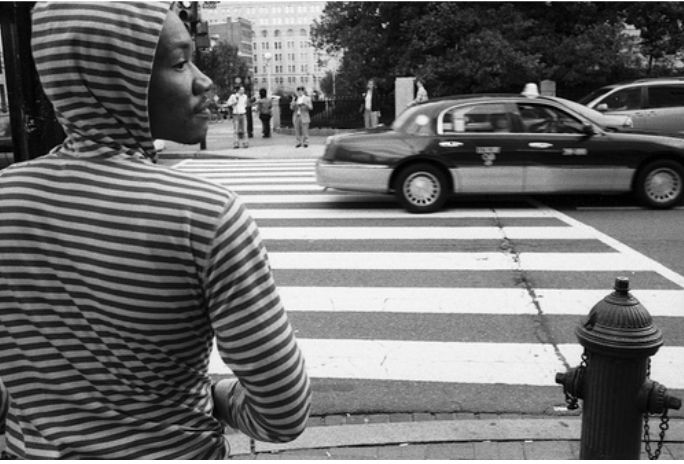}}
\caption{VG image of people at crosswalk.}
\label{fig:people_at_crosswalk}
\end{center}
\end{figure}

In this example, explanations 1--4 are relevant. Example 5, while mentioning the relevant concept of a tennis shoe, does not provide as satisfactory an explanation. Consider another image of a set of people at a crosswalk (Figure~\ref{fig:people_at_crosswalk}) --- for this figure, we get the following results:\\

\noindent {\bf Q/A}: What is across the street? Other people. {\bf Explanation}:\\
\noindent 1. The picture shows: group of people across the street\\
\noindent 2. The picture shows: buildings at the end of the street\\
\noindent 3. The picture shows: people across street near crosswalk\\
\noindent 4. The picture shows: a street lamp\\
\noindent 5. The picture shows: people waiting to cross the street\\

As we can see in this example, explanations 1, 3 and 5 are relevant. The success of these two examples suggests that that getting relevant results in the top 5 explanations.

Our baseline method is to generate a similar explanation but without the attention heatmap guidance from the VQA model. Without the attention heatmap, the sentence generated will still be relevant to the image and question asked, but will not be explaining the decision of the model since the explanation generation is not tied to the model decision process in any way. Thus, this acts as a strong baseline since any relevant sentence to the image shouldn't be mistaken for an explanation of why a model predicted a certain answer for a question. 

We evaluate how the results for this baseline, i.e., in Equation~\ref{eq:score} where the $attentionScore$  is not used. In that case, we expect the results to be of poorer quality. For the tennis example, the following explanations were generated using the attention map:\\

\begin{figure}[hbtp]
\begin{center}
\centerline{\includegraphics[width=0.75\columnwidth]{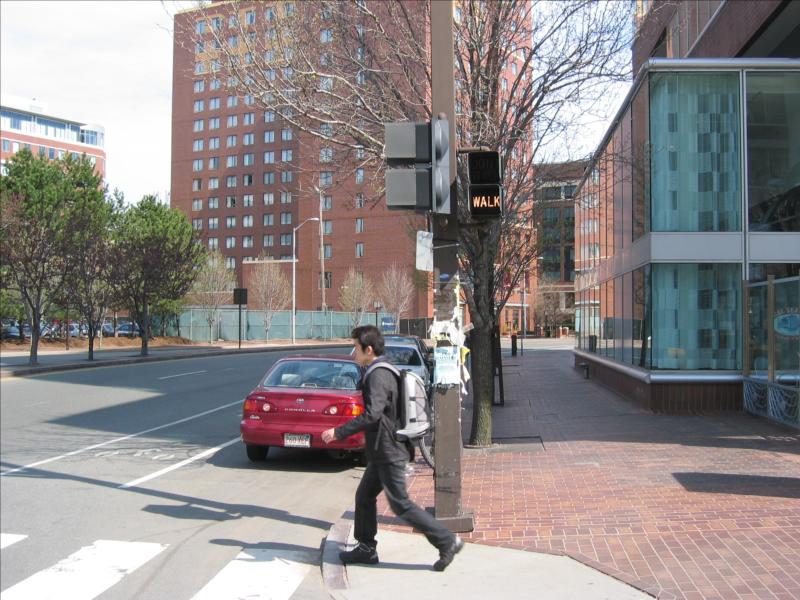}}
\caption{VG image of crosswalk.}
\label{fig:crosswalk}
\end{center}
\end{figure}

\noindent {\bf Q/A}: Question considered: Why is the woman holding a racket? Answer: To hit the ball. {\bf Explanation}:\\
\noindent 1. The picture shows: the tennis racket of the player\\
\noindent 2. The picture shows: a red and silver tennis racket\\
\noindent 3. The picture shows: a woman holding a tennis racket\\
\noindent 4. The picture shows: a woman hitting a tennis ball\\
\noindent 5. The picture shows: a woman hitting a backhand\\

As we can see, explanations 4 and 5 are relevant explanations. In comparison, when we don't use the attention map, we typically find that none of the explanations are quite relevant to the answer:\\

\noindent {\bf Q/A}: Why is the woman holding a racket? To hit the ball. {\bf Explanation}:\\
\noindent 1. The picture shows: a tennis ball \\
\noindent 2. The picture shows: a yellow tennis ball \\
\noindent 3. The picture shows: a small tennis ball \\
\noindent 4. The picture shows: a red and silver tennis racket \\
\noindent 5. The picture shows: the tennis racket of the player \\

To quantify the importance of the attention map in the multi-modal NL explanation generation algorithm, we performed an A/B test with human raters --- the analysis of the ratings, outlined in the next section, validates our assumption that using the multi-modal approach gives us better explanations over using one modality (e.g., linguistic analysis) alone.

We next show some examples of explanations generated by using the story-like algorithm with objects and relations extracted from the scene graph without using region descriptions. In each case, we show the top explanation generated using the story-like explanation generation algorithm, as outlined in Algorithms~\ref{algo2} and~\ref{algo3}:\\

\begin{figure}[hbtp]
\begin{center}
\centerline{\includegraphics[width=0.75\columnwidth]{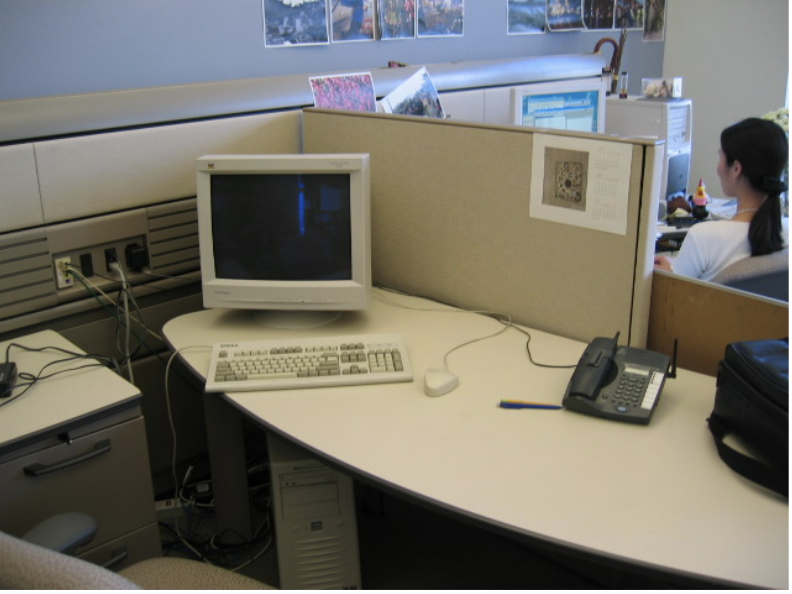}}
\caption{VG image of office.}
\label{fig:office}
\end{center}
\end{figure}

\noindent {\bf Q/A}: Where was this picture taken? At the intersection (Figure~\ref{fig:crosswalk}) \\
{\bf Explanation}: The picture shows crosswalk on road and in front of man, car parked on road, tree next to road, sign next to road, bike next to car, building with window, walk sign.\\

\noindent {\bf Q/A}: Where was this picture taken? In an office (Figure~\ref{fig:office}) \\
{\bf Explanation}: The picture shows keyboard with keys, filing cabinet with drawer, bag on desk, picture on wall, outlet on wall, pen on desk, mouse next to keyboard, filing cabinet with handle, cable on floor, cables on floor.\\

\begin{figure}[hbtp]
\begin{center}
\centerline{\includegraphics[width=0.75\columnwidth]{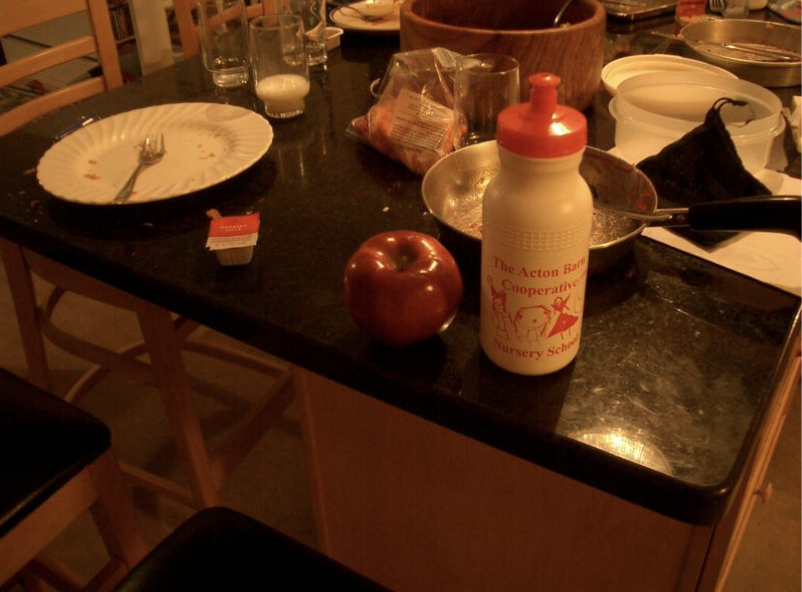}}
\caption{VG image of dining room.}
\label{fig:dining}
\end{center}
\end{figure}

\noindent {\bf Q/A}: Where was this picture taken? In a dining room (Figure~\ref{fig:dining}) \\
{\bf Explanation}: The picture shows chair with leg, food in bag, liquid in glass, fork on plate, bottle with logo, crumb on plate, bag rests on bowl, bar stool, cover, pan.

\section{Experiments}
\label{sec:exp}

We conducted an initial quantitative user-study to demonstrate that human users were satisfied with explanations generated by our multi-modal algorithm.  In this evaluation, the ratings were performed internally within our research group, however, in future experiments, we plan to use Amazon Mechanical Turk. Four internal workers rated about 220 questions --- for each question, explanations were generated using both the multi-modal algorithm and the NL-only approach (as the baseline). In the latter case, visual attention was not used in the explanation generation. Overall, close to 2K explanations were rated. The authors did not know beforehand which of the explanations came from the baseline algorithm vs. our multi-modal approach, so as to not bias ratings in any way. 

We used 3 metrics for evaluation in our small-scale study:

\begin{enumerate}

    \item Explanation score: Each explanation was rated using a relevance score ranging from -5 to +5, where -5 indicates irrelevant explanation, 0 indicates redundant explanation and +5 corresponds to relevant non-redundant explanation. A negative score between 0 and -5 indicates degree of irrelevance, while a positive score between 0 and +5 indicates degree of relevance.
    
    Each question/answer pair was also rated, according to the degree of explainability of the question/answer, on score of 1 to 5 --- 1 indicates that the question/answer pair is difficult to explain (e.g., a question/answer like "Q: What color is the shirt? A: Red"), while 5 indicates that the question is easy to explain (e.g., a question/answer like "Q: Where was this picture taken? A: On the beach").
    
    When an algorithm generates explanations for a question/answer pair, we score each explanation using the relevance score, multiply that score by a position weight (so that explanations in higher positions get higher weight), and finally scale that using the question explainability (so that explanations for more explainable questions are given higher score). 
    
    \item Position of first relevant explanation: Given a set of generated explanations, this metric compares the position of the first relevant explanation.
    
    \item Number of relevant explanations: Given a set of generated explanations, this metric measures the number of relevant explanations in the top-5 generated explanations.

\end{enumerate}

In all 3 cases, the combined multi-modal algorithm outperformed the baseline NL-only approach. The results are summarized in Table~\ref{tab:scores}: considering explanation score, multi-modal was better in $\approx52\%$ cases; based on position on first relevant score, multi-modal was better in $\approx55\%$ cases; while in $\approx54\%$ cases, multi-modal was better than NL-only according to number of relevant explanations. These initial quantitative results demonstrate that humans are more satisfied with our explanations when those explanations came from the multi-modal algorithm that was actually tied to the inference procedure of the visual model.

\begin{table}
\begin{small}
\begin{center}
\begin{tabular}{ |c|c|c|c| } 
 \hline
 Type & Win & Loss & Tie \\ \hline
 Explanation score & 52\% & 28\% & 20\% \\ 
 Position score & 55\% & 30\% & 15\% \\ 
 Number score & 54\% & 24\% & 22\% \\ 
 \hline
\end{tabular}
\caption{Statistics of multi-modal approach compared to NL-only approach (Win $\Rightarrow$ multi-modal wins).}
\label{tab:scores}
\end{center}
\end{small}
\end{table}

\section{Related Work}
\label{sec:rel}

There is a large body of literature on automatic generation of different types of machine explanations, e.g., explanations for recommendation systems~\cite{costa:17}, affordances from images~\cite{chuang:17}, and robotics~\cite{sridharan:16}. Explanation generation has also been explored in the areas of planning~\cite{fox:17}, interactive model debugging~\cite{kulesza:15}, autonomous systems~\cite{langley:17}, mobile robotics~\cite{rosenthal:16}, expert systems~\cite{swartout:91}, tactical behavior modeling~\cite{vanlent:04}, etc.

In this paper, we focus on NL explanations for the visual question answering task, where previous work has been done by Hendricks et al.~\cite{hendricks:2016} and others. A closely-related work to our approach is VQA-X, a method for generating explanation datasets~\cite{park:17}, where the authors propose a multi-modal methodology for simultaneously generating visual and textual explanations.  Another related work for generating NL explanations using a multi-modal approach~\cite{park:18} qualitatively show cases where visual explanation is more insightful than textual explanation (and vice versa), demonstrating that multi-modal explanation models offer significant benefits over uni-modal approaches. Note that both these approaches rely on getting a large corpus of explanations from human annotators for training the models. In our proposed method, we don't need manually generated explanation data, we use already available annotations from scene graphs only.

\begin{figure}[hbtp]
\begin{center}
\centerline{\includegraphics[width=0.6\columnwidth]{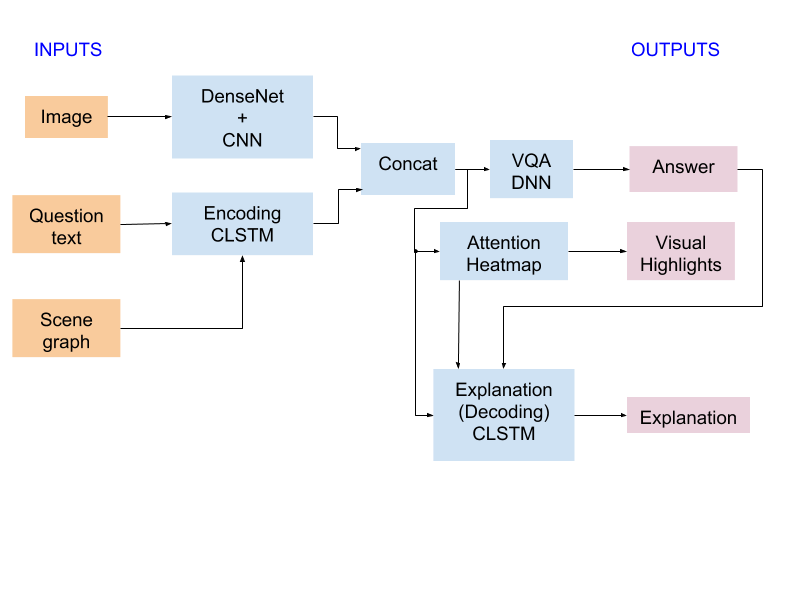}}
\caption{Future Explanation Generation Pipeline.}
\label{fig:final_architecture}
\end{center}
\end{figure}

\section{Conclusions and Future Work}
\label{sec:concl}
In this paper, we presented a multi-modal approach for generating natural language explanations for the visual question answering task, using both visual and linguistic modalities, without collecting any additional data. We also showed empirically how the multi-modal approach gives better explanations than using one modality (e.g., linguistic analysis) alone.

We also plan to look into training effective models for learning to generate explanations while predicting answers given an image, context about the image, and a question. Image context may involve scene graphs and we can use Contextual LSTMs (CLSTMs)~\cite{ghosh:16} to encode this additional information. 
The main benefit of training a pipeline to do explanation generation (as suggested in Figure~\ref{fig:final_architecture}), instead of using hard-coded algorithms (as suggested in Figure~\ref{fig:scoring_pipeline}), is that models have higher flexibility to learn complicated common-sense semantic information, if that is necessary to generate a satisfactory explanation.
 
We would also like to explore some other improvements to our system, namely:
(1) Infer super-categories of relations and objects (e.g., obtained from hypernyms in wordnet), add them as a preface to the explanation list. 
(2) Explore the use of embeddings, e.g., skip-thought vectors~\cite{kiros:15} to find the similarity of entity descriptions (e.g., object attributes, relation phrases or region descriptions) with the salient parts of the question/answer text.
(3) Conducting a study to check if our textual explanations help humans predict VQA accuracy for a given image-question pair. \\

\begin{small}
{\bf Acknowledgments:}
The authors would like to thank Kamran Alipour and Jurgen Schulze for setting up a web demo for our VQA-Interface, and Dr. Ajay Divakaran and Dr. Yi Yao for valuable feedback and suggestions. We would like to thank Dr. Patrick Lincoln for his valuable feedback and support. This research was developed with funding from the Defense Advanced Research Projects Agency (DARPA), under the Explainable Artificial Intelligence (XAI) program. The views, opinions and/or findings expressed are those of the author and should not be interpreted as representing the official views or policies of the Department of Defense or the U.S. Government.
\end{small}

\clearpage
\bibliographystyle{named}
\bibliography{xai-nlp}

\end{document}